\documentclass[letterpaper]{article} 
\usepackage{aaai2026}
\usepackage{times}  
\usepackage{helvet}  
\usepackage{courier}  
\usepackage[hyphens]{url}  
\usepackage{graphicx} 
\usepackage{natbib}  
\usepackage{caption} 
\usepackage{amsmath} 
\usepackage{algorithm}
\usepackage{algorithmic}
\usepackage{amsthm}
\usepackage{mathtools}
\usepackage{amsfonts}
\usepackage{multirow} 
\usepackage{booktabs}
\usepackage{caption}
\usepackage{subcaption}
\usepackage{xcolor}
\usepackage{amsthm,amsmath,amssymb}
\usepackage{mathrsfs}
\usepackage{newfloat}
\usepackage{listings}
\DeclareCaptionStyle{ruled}{labelfont=normalfont,labelsep=colon,strut=off} 
\floatstyle{ruled}
\newfloat{listing}{tb}{lst}{}
\floatname{listing}{Listing}
\title{Personalized Face Super-Resolution with Identity Decoupling and Fitting}
\author{
    Jiarui Yang\textsuperscript{\rm 1 \rm2},
    Hang Guo\textsuperscript{\rm 2},
    Wen Huang\textsuperscript{\rm 2},
    Tao Dai\textsuperscript{\rm 3},
    Shutao Xia\textsuperscript{\rm 2}
}
\affiliations{
    \textsuperscript{\rm 1} Nankai University
    \textsuperscript{\rm 2} Tsinghua University
    \textsuperscript{\rm 2} Shenzhen University
    
}

\usepackage{bibentry}

\begin{document}

\maketitle

\begin{abstract}

In recent years, face super-resolution (FSR) methods have achieved remarkable progress, generally maintaining high image fidelity and identity (ID) consistency under standard settings. However, in extreme degradation scenarios (e.g., scale $> 8\times$), critical attributes and ID information are often severely lost in the input image, making it difficult for conventional models to reconstruct realistic and ID-consistent faces. Existing methods tend to generate hallucinated faces under such conditions, producing restored images lacking authentic ID constraints. To address this challenge, we propose a novel FSR method with Identity Decoupling and Fitting (IDFSR), designed to enhance ID restoration under large scaling factors while mitigating hallucination effects. Our approach involves three key designs: 1) \textbf{Masking} the facial region in the low-resolution (LR) image to eliminate unreliable ID cues; 2) \textbf{Warping} a reference image to align with the LR input, providing style guidance; 3) Leveraging \textbf{ID embeddings} extracted from ground truth (GT) images for fine-grained ID modeling and personalized adaptation. We first pretrain a diffusion-based model to explicitly decouple style and ID by forcing it to reconstruct masked LR face regions using both style and identity embeddings. Subsequently, we freeze most network parameters and perform lightweight fine-tuning of the ID embedding using a small set of target ID images. This embedding encodes fine-grained facial attributes and precise ID information, significantly improving both ID consistency and perceptual quality. Extensive quantitative evaluations and visual comparisons demonstrate that the proposed IDFSR substantially outperforms existing approaches under extreme degradation, particularly achieving superior performance on ID consistency.
\end{abstract}

\section{Introduction}







\begin{figure}[t]
\centering
\includegraphics[width=0.94\columnwidth]{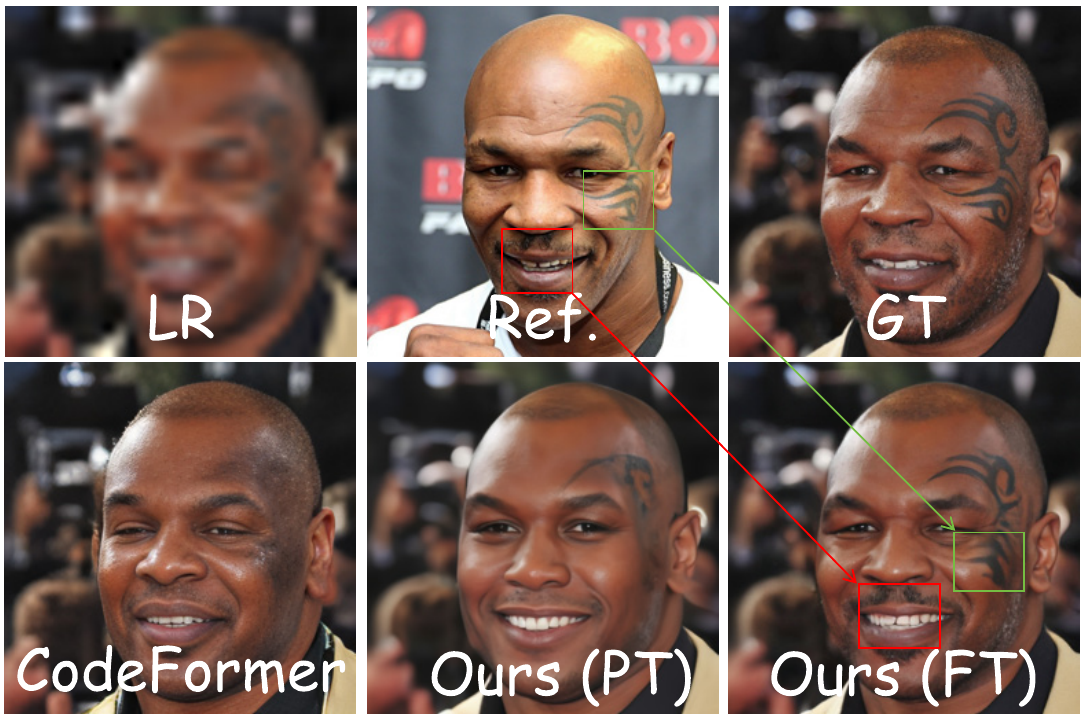} 
\caption{\textbf{Visualization of the pretraining and finetuning result.} Under severe degradation, it is often impossible to reconstruct fine-grained features without relying on reference images. Generalized training can only ensure local ID similarity, whereas personalized fine-tuning demonstrates strong consistency in ID-related attributes.}
\label{fig: intro}
\end{figure}

Face super-resolution (FSR) has significant application value in downstream tasks such as identity (ID) recognition and facial analysis \cite{jiang2021deep, zhang2020msfsr}. In scenarios with strict ID requirements, FSR models must not only enhance image quality but also maintain ID consistency. Existing FSR methods are capable of producing high-fidelity, ID-consistent face images under moderate degradation \cite{chen2018fsrnet, codeformer}. However, under extreme degradation conditions (e.g., scaling factors exceeding $8 \times$), critical ID and attribute information in the input image is often severely lost, turning the reconstruction task into a highly ill-posed problem \cite{alekseev2001analysis}. In the absence of strong prior constraints, the solution space becomes underdetermined, with numerous plausible local optima, making the model prone to generating “hallucinated” results that fail to accurately preserve the target ID. For example, as shown in Fig. \ref{fig: intro}, the conventional method (CodeFormer \cite{codeformer}) may erroneously reconstruct a blurry tattoo region into an unrealistic dark patch.

Reference-based FSR methods \cite{DMDNet, ASFFNet} provide a promising solution to this problem by leveraging high-quality (HQ) semantic information from reference images to restore missing details in low-resolution (LR) inputs. However, single-reference approaches are highly sensitive to pose, lighting, and expression variations, and typically only perform well when the reference and LR images are closely aligned \cite{li2018learning, dogan2019exemplar}. Although multi-reference methods \cite{ASFFNet, MyStyle} attempt to incorporate richer ID features, most existing approaches fail to disentangle ID information between the reference and LR images. This disentanglement is crucial: FSR is a pixel-wise reconstruction task, and the LR image itself contains strong structural priors. During training, models tend to over-rely on the residual ID clues in the LR input, thereby neglecting or underutilizing the complementary information provided by the reference images. As a result, the reconstructed faces may resemble the correct ID to some extent, but fail to accurately reproduce finer attributes and details.

To address these challenges, we propose a diffusion-based two-stage framework for FSR via \textbf{I}D \textbf{D}ecoupling and \textbf{F}itting (IDFSR). IDFSR centers around ID-disentangled pretraining and personalized fine-tuning, aiming to achieve high-fidelity reconstruction with consistent ID and attribute information. Specifically, we introduce three key components to obtain an ID-decoupling representation:

\begin{itemize}
    \item \textbf{Corrupted ID Masking}. We detect the facial region in the LR image and apply masking as a strong conditional constraint. This design preserves background consistency and prevents the model from relying on incomplete or erroneous ID cues during training, thereby enhancing its dependency on and utilization of reference information.
    \item \textbf{Style-Conditioned Embedding}. We spatially align the reference face region to the LR face and extract a style embedding via a style encoder. This embedding is integrated into the diffusion model through feature fusion, providing coarse-grained appearance and motion guidance from the reference image.
    \item \textbf{Ground Truth ID Embedding and ID Fitting}. During pretraining, we extract ID embeddings from GT images using a well-trained ID encoder and integrate them into the model via cross-attention. In the fine-tuning, we freeze all model parameters and optimize the ID embeddings using a few samples from the same ID to achieve generalized and precise decoupling representation.
\end{itemize}

As shown in Fig. 1, the pretraining phase yields reconstructions with basic ID similarity, while ID Fitting significantly enhances attribute fidelity and fine-grained details. Extensive experiments across multiple benchmark datasets validate the effectiveness of each proposed component and the framework’s ability to suppress hallucination artifacts. Furthermore, comprehensive quantitative comparisons with various SOTA methods show that our approach achieves a significant performance improvement of 20\%, demonstrating the clear superiority of IDFSR in both reconstruction quality and ID consistency.

\section{Related Works}

\textbf{Single-Image Face Super-Resolution.}
Different from natural image super-resolution \cite{zhang2018rcan,dai2019san,guo2024mambair, guo2025mambairv2,dai2024freqformer}, face super-resolution (FSR) \cite{tomar2023comprehensive} poses unique challenges due to the need for accurate restoration of fine-grained facial details, identity consistency, and structural priors. Early methods on FSR predominantly relied on direct regression in pixel space; however, empirical studies have shown that such approaches tend to produce overly smoothed results \cite{huang2017wavelet,chen2018fsrnet}. To improve perceptual quality, generative models have been introduced, leveraging prior distributions or explicit dictionaries to enhance the naturalness of reconstructed images \cite{he2022gcfsr, pgdiff}. Nonetheless, due to the inherent ambiguity in generative modeling, these methods still face challenges in maintaining pixel-level consistency and ID fidelity. To balance consistency and diversity, recent studies have incorporated conditional generation mechanisms, using facial priors such as landmarks, segmentation maps, and 3D structures as auxiliary constraints to enforce geometric consistency \cite{yang2025diffusion, bulat2018super}. While these approaches have improved geometric reconstruction to some extent, ID preservation remains limited. To address this, ID recognition losses have been employed to enhance ID fidelity \cite{chen2020identity}. However, under large degradation factors, ID cues in LR inputs are severely diminished, making models prone to generating "hallucinated" identities.

\textbf{Reference-Based Face Super-resolution.}
To alleviate the severe information loss under high degradation, reference-based FSR methods have been extensively explored. GFRNet \cite{li2018learning} proposed a dual-network architecture—WarpNet aligns the reference image to the LR target via optical flow, while RecNet performs reconstruction. Landmark loss and total variation regularization are adopted to enhance alignment accuracy. However, large pose or expression differences still lead to misalignment issues. GWAINet \cite{dogan2019exemplar} later introduced ID loss to improve ID consistency but remained heavily dependent on precise alignment. ReFine \cite{chong2025copy} improved alignment using a fine-tuned landmark detector and employed spatial minimality and cycle consistency losses to guide attribute transfer. ASFFNet \cite{ASFFNet} selected suitable references via landmark similarity and aligned them in spatial and illumination domains using MLS and AdaIN, followed by multi-stage feature fusion. DMDNet \cite{DMDNet} used a dual-memory structure to separately store structural and ID features, enabling adaptive fusion across references. MyStyle \cite{MyStyle} and similar methods construct identity and attribute sets from multiple references.  Recently, MGFR \cite{tao2024overcoming} utilized diffusion models to fuse text, reference images, and ID cues via dual-control adapters and two-stage training for controllable face restoration.

\begin{figure*}[t]
\centering
\includegraphics[width=2\columnwidth]{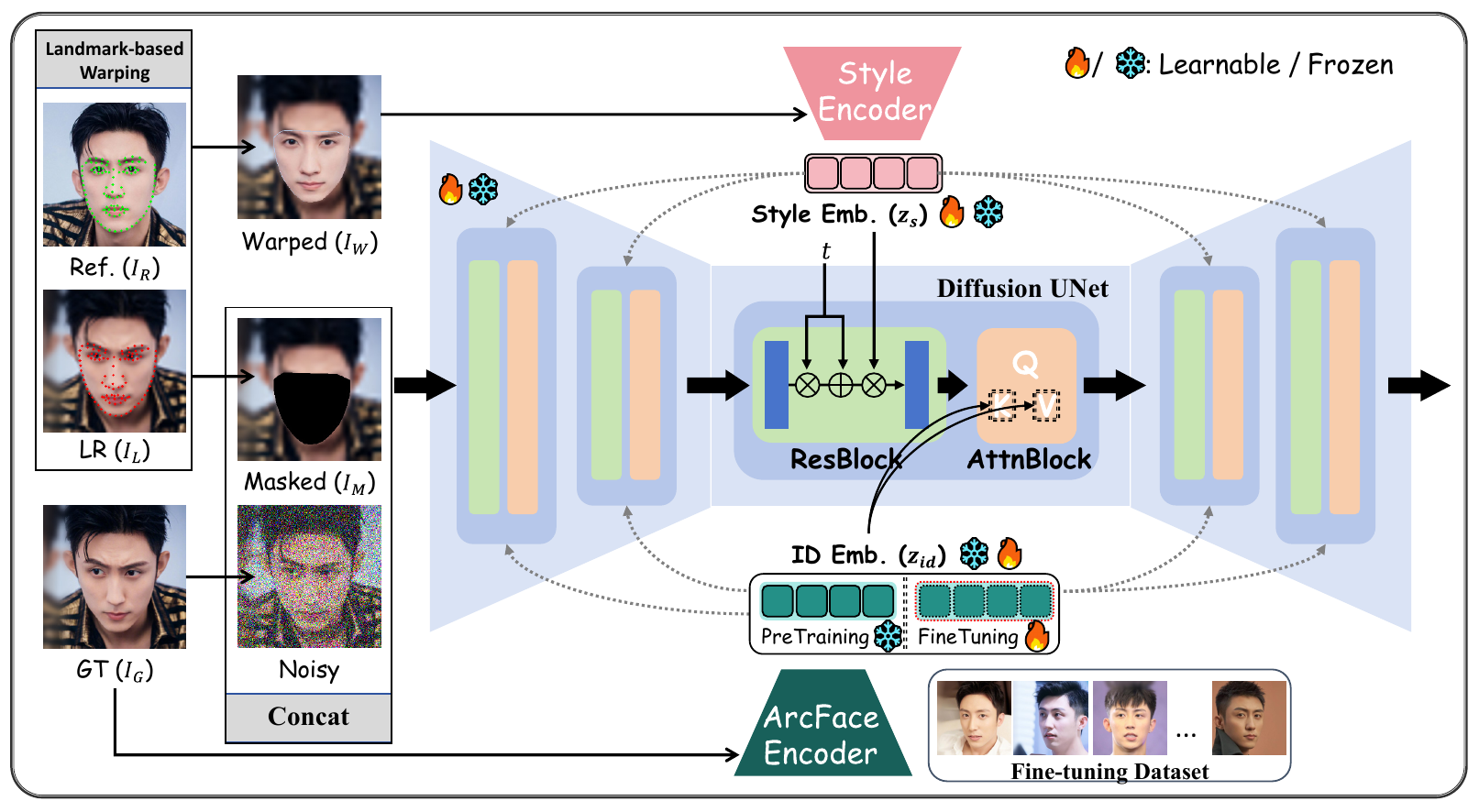} 
\caption{A schematic diagram of single-step diffusion in IDFSR, including input preprocessing and the overall model architecture.}
\label{fig: framework}
\end{figure*}

\section{Methodology}
\subsection{Overview}
Fig. \ref{fig: framework} illustrates the overall architecture of IDFSR. Given a LR image $I_L$, its corresponding GT image $I_G$, and a same-identity reference image $I_R$, IDFSR consists of a parameterized network $\theta$, a style encoder $\mathcal{E}_\mathrm{s}$, and a pretrained ArcFace ID encoder $\mathcal{E}_{\mathrm{id}}$ \cite{deng2019arcface}. Prior to training, we perform two preprocessing steps:
\begin{itemize}
    \item A face landmark detector is used to mask the facial region of $I_L$, producing the masked image $I_M$.
    \item A landmark-based warping is applied to map the facial region of $I_R$ onto that of $I_L$, generating the warped image $I_W$.
\end{itemize}
In the pretraining stage, we adopt Denoising Diffusion Probabilistic Model (DDPM \cite{ddpm}) to model the pixel space, using the style embedding $z_{\text{s}} = \mathcal{E}_{\mathrm{s}}(I_W)$ and ID embedding $z_{\text{id}} = \mathcal{E}_{\mathrm{id}}(I_G)$ as conditions. $I_M$ is concatenated with the noise as input. The simplified loss function at diffusion timestep $t$ is defined as:

\begin{equation}
\mathcal{L}_{\text{sim}} = \mathbb{E}_{x_0, \epsilon \mid z_{\text{s}}, z_{\text{id}}, I_M} 
\left\{ \left\| \epsilon - \epsilon_\theta([x_t, I_M], t, z_{\text{s}}, z_{\text{id}}) \right\|^2 \right\}.
\end{equation}
where [·] denotes the concatenation operation, and $x_t$ represents the noisy of the GT image at time step $t$. In the finetuning stage, we freeze the diffusion network $\theta$ and the style encoder $\mathcal{E}_\mathrm{s}$, and replace $z_{\text{id}}$ with a trainable embedding vector. This vector is then optimized using a small number of same-identity samples, under the same training objective, enabling personalized and accurate ID control.

\begin{figure}[t]
\centering
\includegraphics[width=0.99\columnwidth]{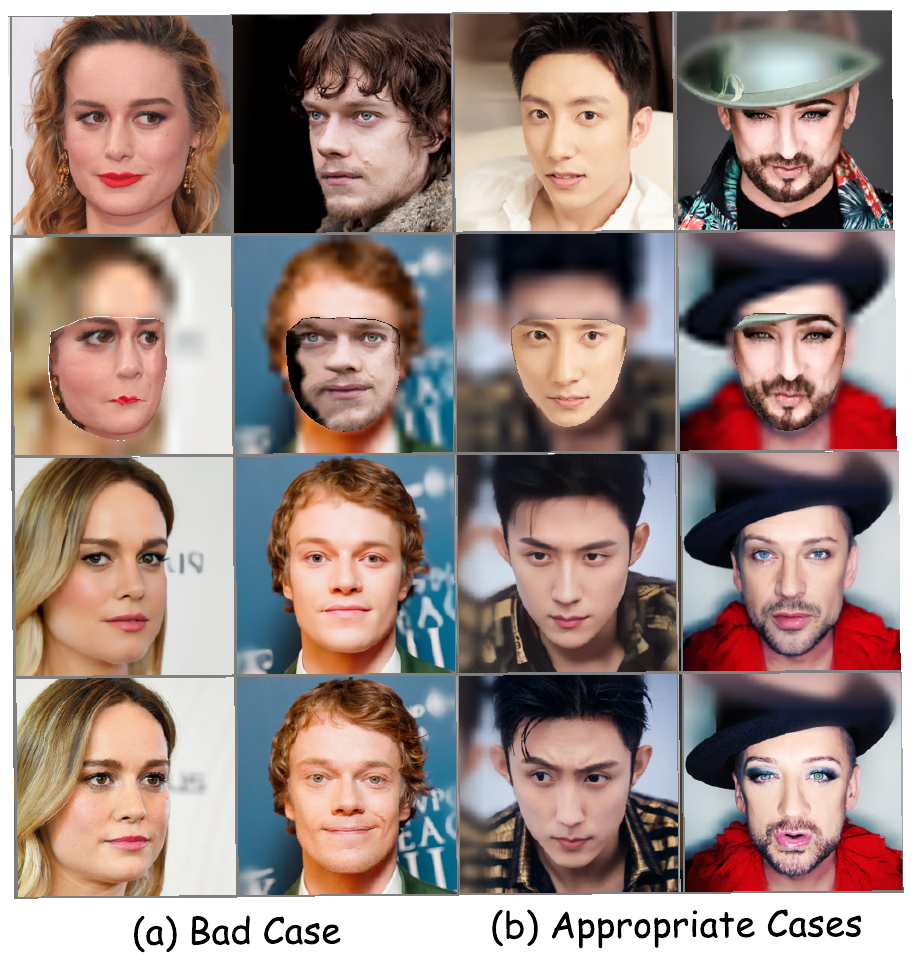} 
\caption{\textbf{Warping Analysis}. From top to bottom: reference images, warped images, SR images, and GT images.}
\label{fig: warping}
\end{figure}

\subsection{Landmark Detection and Warping}
We use RetinaFace \cite{deng2020retinaface} to extract facial landmarks from HQ images and finetune the detector to better accommodate LR images. The facial region of the reference image $I_R$ is then aligned to that of the LR image $I_L$ via affine transformation based on the detected landmarks, resulting in a warped image $I_W$. This pseudo-alignment introduces coarse spatial correspondence between the reference and the LR image. We provide the pseudocode of the warping process in the Appendix.

However, due to variations in pose or inaccuracies in landmark prediction, such warping is often imperfect and may introduce local distortions or mismatches. Interestingly, we find that this imperfection does not hinder the model's performance—instead, it acts as a form of natural \textbf{data augmentation}. The diversity in warped faces encourages the model to avoid over-reliance on precise spatial correspondence and instead to attend to global ID semantics. This drives the diffusion network to rely more heavily on the ID embedding $z_{\text{id}}$ for reconstructing missing facial regions, rather than naively copying textures from the warped reference. As a result, our model becomes inherently robust to variations in reference quality and alignment. As shown in Fig.~\ref{fig: warping}, IDFSR still generating plausible and ID-consistent faces even under severe misalignment.

\begin{figure*}[t]
\centering
\includegraphics[width=2.12\columnwidth]{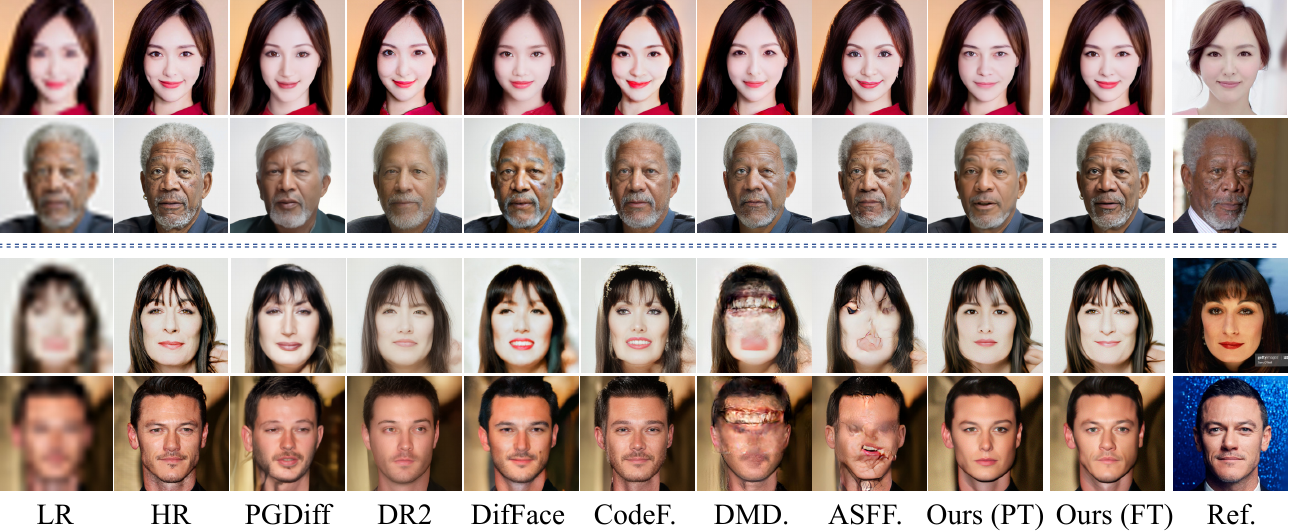} 
\caption{Visualization results of different methods under upsampling scales of 8 and 16, separated by double dashed lines. For reference-based methods—DMDNet, ASFFNet, and our IDFSR—a single ID is randomly selected as the reference image. FT and PT refer to finetuning and pretraining methods, respectively. Please zoom in for a better viewing.}
\label{fig: vis_main}
\end{figure*}

\begin{table*}[t]
  \centering
  \scalebox{0.75}{
      \begin{tabular}{l|c c c c c c c|c c c c c c c}
        \toprule[1.5pt]
        & \multicolumn{7}{c|}{$\times8$} & \multicolumn{7}{c}{$\times16$} \\
         Methods & PSNR$\uparrow$ & SSIM$\uparrow$ & LPIPS$\downarrow$ & FID$\downarrow$  & CLIPIQA$\uparrow$ & MUSIQ$\uparrow$ & IDS$\downarrow$ &PSNR$\uparrow$ & SSIM$\uparrow$ & LPIPS$\downarrow$ & FID$\downarrow$  & CLIPIQA$\uparrow$ & MUSIQ$\uparrow$ & IDS$\downarrow$\\
        \midrule[1.5pt]
         CodeFormer & \textcolor{blue}{24.42} & 0.7147 & \textcolor{blue}{0.1489} & 31.88 & \textcolor{blue}{0.6873} & 68.98 & 0.3865 & 20.72 & 0.5947 & 0.2379 & 43.89 & \textcolor{blue}{0.6834} & 67.85 & 0.6673 \\
         PGDiff & 22.73 & 0.6892 & 0.2223 & 46.05 & 0.5390 & 56.10 & 0.6080 & 20.47 & 0.6202 & 0.2727 & 55.70 & 0.5008 & 52.19 & 0.7355 \\
         DR2 & 23.83 & 0.7067 & 0.1998 & 44.46 & 0.6307 & 63.35 & 0.5459 & 22.35 & 0.6356 & 0.2481 & 50.84 & 0.6102 & 62.60 & 0.7082 \\
         DifFace & 24.41 & \textcolor{blue}{0.7255} & 0.1694 & 32.94 & 0.6132 & 60.54 & 0.3815 & 22.23 & 0.6737 & 0.2195 & 38.65 & 0.5957 & 57.84 & 0.5948 \\
         ASFFNet & 21.82 & 0.6624 & 0.1764 & \textcolor{red}{26.25} & 0.6158 & 65.81 & \textcolor{blue}{0.3169} & 18.23 & 0.5732 & 0.2457 & 40.64 & 0.5827 & 63.72 & 0.7955 \\
         DMDNet & 22.36 & 0.6984 & 0.1733 & 29.76 & 0.6570 & 68.28 & 0.4169 & 17.38 & 0.5590 & 0.2497 & 46.34 & 0.6051 & 64.05 & 0.8291 \\
         Ours (PT) & 24.29 & 0.6996 & 0.1554 & 27.89 & 0.6798 & \textcolor{blue}{69.30} & 0.3173 & \textcolor{blue}{23.32} & \textcolor{blue}{0.6863} & \textcolor{blue}{0.2062} & \textcolor{blue}{38.58} & 0.6723 & \textcolor{blue}{68.56}& \textcolor{blue}{0.5227} \\
         Ours (FT) & \textcolor{red}{28.85} & \textcolor{red}{0.7604} & \textcolor{red}{0.1031} & \textcolor{blue}{26.69} & \textcolor{red}{0.7092} & \textcolor{red}{73.73} & \textcolor{red}{0.2242} & \textcolor{red}{24.09} & \textcolor{red}{0.7158} & \textcolor{red}{0.1845} & \textcolor{red}{35.50} & \textcolor{red}{0.6992} & \textcolor{red}{71.83} & \textcolor{red}{0.3625} \\
        \bottomrule[1.5pt]
      \end{tabular}
      }
  \caption{Quantitative comparison on the CelebRef-HQ dataset with upsampling scales of 8$\times$ and 16$\times$. FT and PT refer to finetuning and pretraining methods, respectively. \textcolor{red}{\textbf{Red}} and \textcolor{blue}{blue} indicates the best and the second best.}\label{Tab:1}
\end{table*}

\subsection{Pretraining and Finetuning}
Our diffusion backbone follows the U-Net architecture used in Guided Diffusion \cite{dhariwal2021diffusion}, consisting of residual blocks, attention modules, and upsampling/downsampling layers. Our innovative conditional injection method effectively achieves ID decoupling and fitting modeling.

The style encoder adopts the encoder part of the diffusion U-Net. Inspired by DiffAE \cite{preechakul2022diffusion}, we introduce adaptive group normalization (AdaGN \cite{wu2018group}) to inject style conditions. For a feature map $h \in \mathbb{R}^{c \times h \times w}$, the AdaGN is formulated as:

\begin{equation}
\text{AdaGN}(h, t, z_s) = z_f \cdot \left( t_s\text{GroupNorm}(h) + t_b \right),
\end{equation}
where $z_f \in \mathbb{R}^c$ is obtained via an affine transformation applied to the style embedding $z_{\text{s}}$:
\begin{equation}
z_f = \text{MLP}_{\text{style}}(z_{\text{s}}),
\end{equation}

and $(t_s, t_b) \in \mathbb{R}^{2 \times c}$ are obtained by applying an MLP to the sinusoidal position embedding $\psi(t)$:
\begin{equation}
(t_s, t_b) = \text{MLP}_{\text{time}}(\psi(t)),
\end{equation}

In the pretraining, we leverage a pretrained ArcFace encoder \cite{deng2019arcface} to extract ID embeddings from the GT images. These ID embeddings are injected into the attention modules via cross-attention mechanisms to provide fine-grained ID priors. The diffusion network and the style encoder are trained from scratch, initialized with DiffAE weights pretrained on the FFHQ dataset \cite{ffhq}. In the finetuning stage, we freeze both the U-Net and the style encoder, and solely optimize a learnable ID embedding initialized from a reference image. Note that the ArcFace encoder is not required during finetuning. Both stages share the same training hyperparameters: a maximum diffusion step of $T = 1000$, a linear $\beta$ schedule, and DDIM sampling with 20 steps during inference.

\section{Experiments}
\subsection{Experimental Setting}
aOur comparison experiments include both general and personalized ones. In the pretraining phase, ID embeddings from reference images are used to achieve general FSR, while in the fine-tuning phase, we focus more on personalized FSR for the same ID. The pretraining strategy is based on the DiffAE model, which was trained for 2 days on four 3090 GPUs, followed by 20 minutes of fine-tuning on a single 3090 GPU with the same ID samples. Detailed implementations are provided in the Appendix. 
\\
\textbf{Datasets}. We uniformly sample 10\% of the IDs from CelebRef-HQ, with the remaining 90\% used for pretraining. Each sampled ID is equally divided into a fine-tuning set and a test set. We filter out certain identities that may contain only a small number of images and additionally select several representative identities to better evaluate ID fitting. Consequently, the training set contains approximately 900 IDs and 9,000 images, while the test set contains 56 IDs and 586 images. Additionally, CASIA-WebFace and the video dataset CelebText are used for generalization evaluation after quality filtering. See the Appendix for further details. 
\\
\textbf{Metrics and Baselines.} For evaluation, we adopt metrics, including PSNR, SSIM, LPIPS \cite{zhang2018unreasonable}, FID \cite{heusel2017gans}, CLIPIQA \cite{wang2023exploring}, MUSIQ \cite{ke2021musiq}, and ID similarity (IDS) measured by DeepFace \cite{serengil2020lightface}. We compare our method against SOTA reference-free and reference-based approaches, including CodeFormer \cite{codeformer}, DR2 \cite{dr2}, DifFace \cite{difface}, PGDiff \cite{pgdiff}, DMDNet \cite{DMDNet}, and ASFFNet \cite{ASFFNet}.

\subsection{Qualitative Comparisons}
As shown in Fig. \ref{fig: vis_main}, we present a qualitative comparison of current SOTA FSR methods. Under moderate degradation ($8 \times$), non-reference methods are able to achieve high-fidelity image reconstruction, but they still exhibit significant discrepancies in facial consistency and fine-grained detail recovery. Unconditional generative models, such as PGDiff and DR2, tend to generate facial hallucinations, i.e., unrealistic or erroneous facial features. In contrast, conditional generative methods with discriminative constraints, such as DifFace, perform better in pixel-level consistency but still struggle to preserve fine ID features. This limitation primarily arises from the tendency of these methods to learn local average during training, which diminishes their ability to model fine-grained features. Our pretrained model exhibits similar characteristics to some extent, but by incorporating a reference image, it outperforms non-reference methods in alleviating facial hallucinations and improving ID consistency. Meanwhile, SOTA reference-based methods like ASFFNet maintain a better balance between ID consistency and attribute recovery at $8 \times$ degradation. However, their performance significantly deteriorates under extreme degradation (such as $16 \times$), with their effectiveness heavily reliant on precise matching between the LR and reference images. In contrast, our proposed personalized fine-tuning mechanism significantly enhances the model's ability to preserve ID features and attribute details, demonstrating greater robustness under extreme low-quality conditions. Under the $16 \times$ setting, IDFSR achieves notable advantages in both ID consistency and visual fidelity.

\subsection{Quantitative Comparisons}
In fact, there exists a discrepancy in embeddings between the training and testing stages of the pretraining method. Specifically, GT ID embeddings are used during pretraining, whereas reference ID embeddings are employed during testing. This mismatch may lead to model performance being influenced by the fine-grained details of the reference image, potentially resulting in misalignment of attribute features. We will delve deeper into this potential editing effect in the next Section. 

Notably, under moderate degradation (as shown in Table \ref{Tab:1}
), the pretrained model does not achieve optimal performance in pixel-level consistency metrics such as PSNR, SSIM, and LPIPS. However, under severe degradation (e.g., $16 \times$), attribute transfer significantly enhances performance, enabling the pretrained model to achieve SOTA results in pixel-level consistency. Further analysis reveals that perceptual quality metrics—including FID, CLIPIQA, and MUSIQ—also demonstrate competitive performance, validating the effectiveness of the pretraining paradigm. As illustrated in Table \ref{Tab:1}, our personalized fine-tuning strategy achieves remarkable SOTA results across pixel-level consistency, perceptual quality, and ID consistency. Specifically, fine-tuning improves pixel-level consistency metrics by approximately 15\%, significantly boosts ID consistency, and surpasses other methods while maintaining high visual fidelity.

\begin{table}[t]
  \centering
  \scalebox{0.77}{
      \begin{tabular}{l|c c c c c}
        \toprule[1.5pt]
         Methods & IDV $\uparrow$ & $\text{ACC}_\text{Gen}$ $\uparrow$ & $\text{ACC}_\text{Emo}$ $\uparrow$ & $\text{ACC}_\text{Ra}$ $\uparrow$ & $\text{Diff}_\text{Age}$ $\downarrow$ \\
        \midrule[1.2pt]
         CodeFormer & 50.3\% & 95.2\% & 64.3\% & 64.8\% & $4.6 \pm 4.3$ \\
         PGDiff & 32.9\% & 88.7\% & 51.0\% & 68.8\% & $5.2 \pm 4.6$ \\
         DR2 & 38.9\% & 91.9\% & 57.6\% & 71.5\% & $5.2 \pm 5.1$ \\
         DiffFace & 73.0\% & 95.4\% & 67.4\% & 77.3\% & $4.6 \pm 4.4$ \\
         ASFFNet & 21.4\% & 72.1\% & 39.6\% & 45.2\% & $7.5 \pm 7.2$ \\
         DMDNet & 18.7\% & 66.3\% & 36.8\% & 42.5\% & $7.6 \pm 7.3$ \\
        \midrule
         \textbf{Ours (FT)} & \textbf{89.6}\% & \textbf{98.7}\% & \textbf{66.2}\% & \textbf{96.6}\% & \textbf{3.3} $\pm$ \textbf{1.7} \\
        \bottomrule[1.5pt]
      \end{tabular}
  }
  \caption{The qualitative comparison of face ID verification (IDV) and attribute consistency, including gender accuracy ($\text{ACC}_\text{Gen}$), emotion accuracy ($\text{ACC}_\text{Emo}$), race accuracy ($\text{ACC}_\text{Ra}$), and age difference ($\text{Diff}_\text{Age}$).}\label{Tab:2}
\end{table}

\begin{figure*}[t]
\centering
\includegraphics[width=2.0\columnwidth]{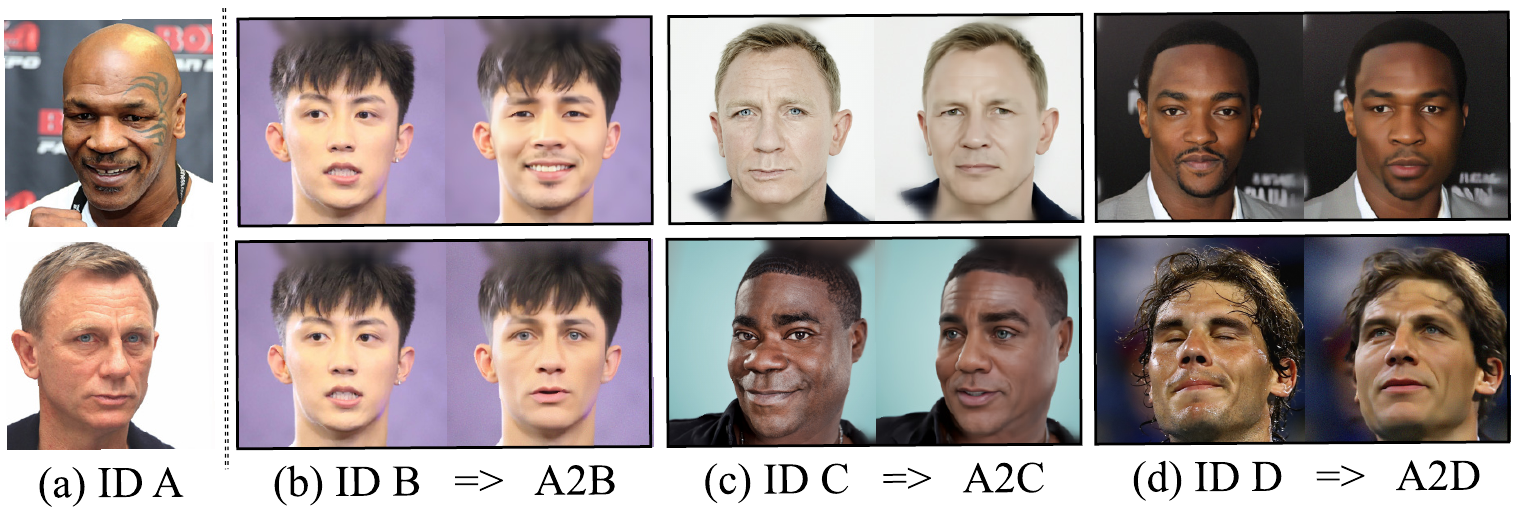} 
\caption{\textbf{Visualization of cross-ID attribute transfer}. We first fit the embedding on ID A, then perform FSR using the LR and style ID from another ID.}
\label{fig: cross}
\end{figure*}

\begin{figure}[t]
\centering
\includegraphics[width=1\columnwidth]{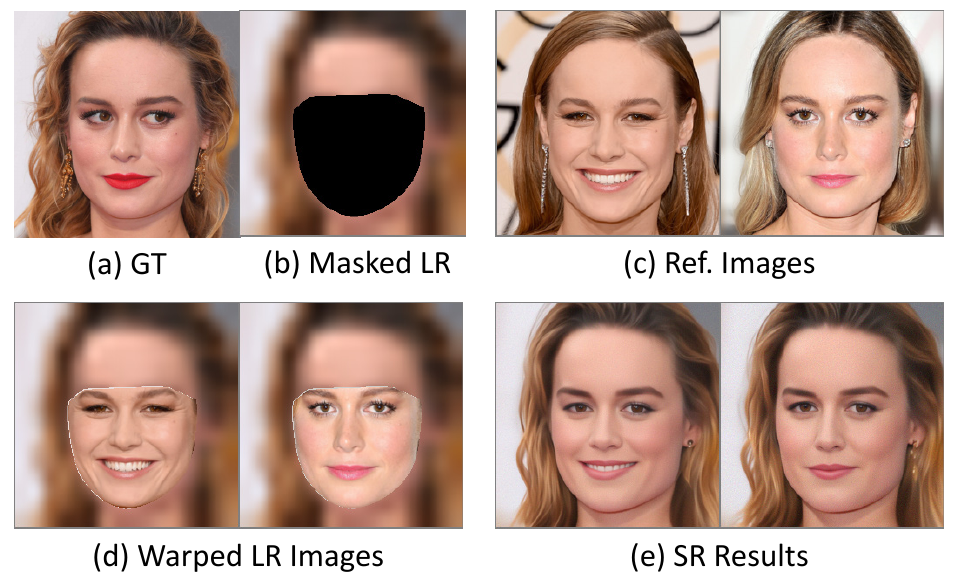} %
\caption{\textbf{Latent Editing Effects of Style Embedding.} Using different styles of reference images usually only changes the action features, while the attributes and background features have a high degree of consistency.}
\label{fig: eidt}
\vspace{-1cm}
\end{figure}

\subsection{Face Verification and Attribute Analysis}
We further perform extensive evaluations on fine-grained attribute consistency to verify the semantic reconstruction ability of our method. Specifically, we use the DeepFace \cite{jiang2021deep} analysis toolkit to conduct face verification and attribute prediction between the GT and the corresponding SR images. Face verification is based on ID similarity, where a default threshold determines whether the GT and SR images belong to the same person. The attribute prediction includes four dimensions: age, gender, emotion, and race.

We quantitatively compare different methods using recognition accuracy under a $16 \times$ scale setting, as shown in Table \ref{Tab:2}. IDFSR shows a clear advantage in ID verification accuracy. In comparison, many existing methods achieve less than 50\% accuracy, indicating significant identity distortion and demonstrating the robustness of our approach under severe degradation. While most methods maintain reasonable gender consistency, emotion and race consistency remain challenging. Our method achieves 96\% accuracy in race prediction, which benefits from the use of personalized ID embeddings However, the accuracy for emotion prediction is notably lower. This is mainly because our method does not incorporate strong priors from the LR image space. Instead, it generates faces based on the reference image and customized embeddings, which can affect the expression. We further discuss this issue in the next Section.

\section{Ablation Study}

\subsection{ID Decoupling and Generalization Verification}
We design a cross-ID experiment to evaluate the fine-grained modeling capability of the learned ID representation. Specifically, we fit an ID embedding on images of a subject (ID A), and then apply this embedding to restore images of a different subject (ID B). During testing, the embedding of ID A is used in conjunction with LR and reference images from other identities. As shown in Fig. \ref{fig: cross}, despite the mismatch in ID, the fitted ID A embedding can still effectively modify the target facial attributes, while preserving the background and style. This observation suggests that the ID embedding is decoupled from the LR input and style encoding, indicating that it is learned independently of identity-aligned pixel supervision or image warping processes. Moreover, the success of the cross-ID experiment further demonstrates the strong generalization capability of the fitted ID representation.

\subsection{Latent Editing Effects of Style Embedding}
The style embedding $z_s$ serves as the only explicitly designed pathway in our framework for capturing spatial information of the face, although the warping operation is not always perfectly aligned. In Methodology, we have discussed the impact of warping alignment; here, we further explore the potential editing effect of style embedding in the context of FSR.

As shown in Fig. \ref{fig: eidt}, we select three images of the same ID with variations in lighting, makeup, and facial expressions. Two of these, which exhibit large differences in facial motion, are used as reference images under $16 \times$ downsampling. It can be clearly observed that the SR outputs exhibit a noticeable resemblance in facial expression to the reference images. Despite certain misalignments at the pixel level, our method achieves high identity consistency and visual fidelity. In contrast, most existing approaches prioritize pixel-level alignment, which restricts the flexibility of ID generation. Notably, aside from variations in facial expressions, the differences in background and fine details are minimal. This further validates the effectiveness of our key design in decoupling ID features.

\begin{figure}[t]
\centering
\includegraphics[width=1\columnwidth]{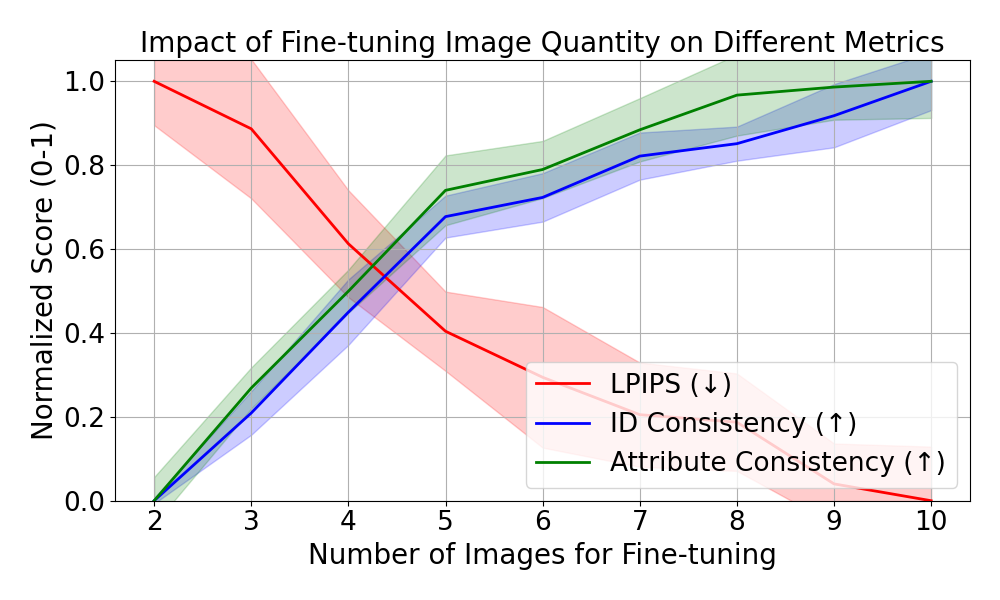} 
\caption{\textbf{Effect of Reference Image Quantity on Finetuning Performance.} Results show that performance generally improves with more references, with diminishing returns beyond five images, where performance and stability reach a balance.}
\label{fig: ft_set}
\end{figure}

\subsection{The Impact of Reference Image Quantity}
To investigate the impact of the number of reference images on fine-tuning performance, we select samples from the CelebHQ-Ref test set, where each ID contains more than 10 images. Specifically, we gradually increase the number of reference images used for fine-tuning, ranging from 2 to 10, and evaluate the performance using three normalized metrics: pixel-level, ID and attribute consistency. As shown in Fig. \ref{fig: ft_set}, we report the average values and confidence intervals of these metrics across 10 identities.

The overall trend shows that increasing the number of reference images helps improve the model’s consistency across all evaluation metrics. In the early stages, introducing a small number of images brings limited improvement, which may be due to the pretrained model already possessing a certain level of single-image reference capability. The additional images provide limited information in terms of attribute diversity or fine-detail supplementation, especially when there are variations in image quality or pose. Notably, when the number of reference images reaches around five, the performance and stability tend to reach an optimal balance.

\subsection{Component Analysis}
\begin{figure}[t]
\centering
\includegraphics[width=1\columnwidth]{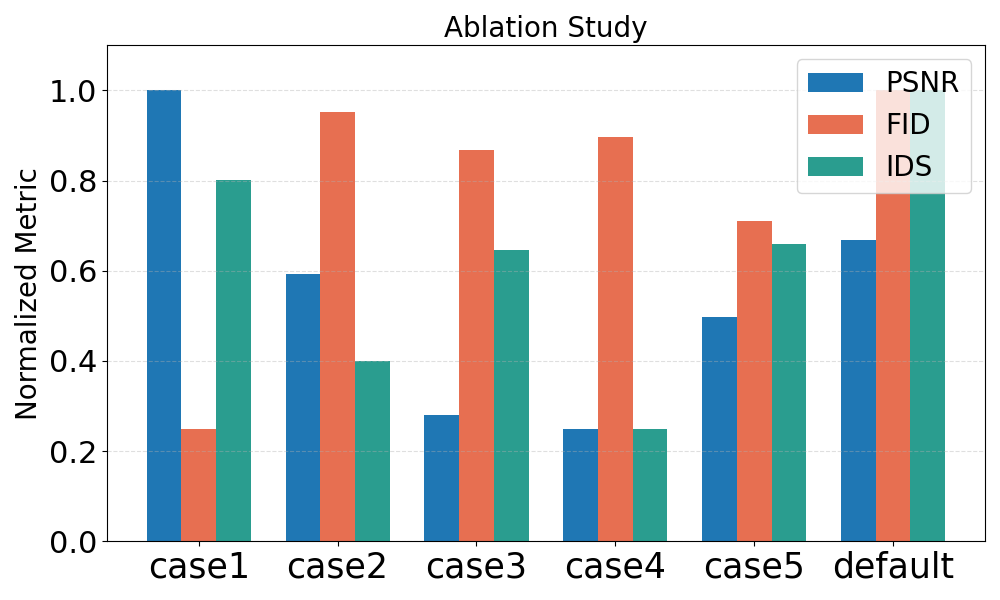} 
\caption{\textbf{Ablation Study on Model Components}. Since the metrics have different scales, we normalize them to better illustrate the impact of each component on various performance measures.}
\label{fig: ab_study}
\end{figure}

Starting from an unconditional diffusion model, we progressively introduce key components to systematically analyze the contribution of each element to the final performance, as illustrated in Fig. \ref{fig: ab_study}. The most basic yet effective diffusion-based super-resolution strategy concatenates the low-resolution (LR) image with the noise input at each timestep (case1), as exemplified by SR3 \cite{saharia2022image} and SRDiff \cite{li2022srdiff}. This strategy imposes strong constraints that improve generation consistency, but suffers from reduced training efficiency and reconstruction fidelity under severe degradation.

In case2, the LR image is masked before concatenation, effectively transforming the task into an inpainting problem. This introduces greater generative uncertainty. Nevertheless, such strong prior constraints remain crucial for low-level vision tasks. Compared to methods relying on feature-based conditioning or attention mechanisms—such as DiffAE and Stable Diffusion \cite{rombach2022high} (case3)—this approach offers superior pixel-level consistency but performs worse in terms of perceptual quality. As shown in Fig. \ref{fig: ab_study}, case1 achieves the best consistency scores but significantly degrades perceptual quality, while case2 exhibits the opposite trend. Case3 yields moderate performance across all metrics, indicating that it tends to produce structurally plausible but not precisely aligned results.

When the ID embedding is removed under the default setting (case4), we observe that the model tends to generate ID-inconsistent faces due to misalignments introduced by the warping operation, resulting in decreased ID consistency. Removing the style embedding (case5) leads to reconstructions that often lack realistic ID traits, suggesting that ID and style embeddings play complementary roles in preserving ID fidelity. The default configuration effectively integrates the advantages of each component: the masking operation provides a strong background constraint prior, while the ID and style embeddings enable disentangled modeling and accurate fitting of ID and facial attributes. Consequently, the model achieves strong performance in both ID consistency and attribute preservation.

\section{Conclusion}
In this paper, we propose IDFSR, a two-stage diffusion-based face super-resolution framework. During the pretraining stage, we achieve facial disentanglement through three key designs, while the finetuning stage enables the fitting of personalized attributes. IDFSR is both generalizable and customizable, demonstrating impressive identity and attribute consistency in both quantitative and qualitative evaluations. Furthermore, extensive ablation studies validate the disentanglement and editability capabilities of our method.

\bibliography{main}

\begin{thebibliography}{41}
\providecommand{\natexlab}[1]{#1}

\bibitem[{Alekseev and Navon(2001)}]{alekseev2001analysis}
Alekseev, A.~K.; and Navon, I.~M. 2001.
\newblock The analysis of an ill-posed problem using multi-scale resolution and second-order adjoint techniques.
\newblock \emph{Computer Methods in Applied Mechanics and Engineering}, 190(15-17): 1937--1953.

\bibitem[{Bulat and Tzimiropoulos(2018)}]{bulat2018super}
Bulat, A.; and Tzimiropoulos, G. 2018.
\newblock Super-fan: Integrated facial landmark localization and super-resolution of real-world low resolution faces in arbitrary poses with gans.
\newblock In \emph{Proceedings of the IEEE conference on computer vision and pattern recognition}, 109--117.

\bibitem[{Chen et~al.(2020)Chen, Chen, Wang, Liang, and Lin}]{chen2020identity}
Chen, J.; Chen, J.; Wang, Z.; Liang, C.; and Lin, C.-W. 2020.
\newblock Identity-aware face super-resolution for low-resolution face recognition.
\newblock \emph{IEEE Signal Processing Letters}, 27: 645--649.

\bibitem[{Chen et~al.(2018)Chen, Tai, Liu, Shen, and Yang}]{chen2018fsrnet}
Chen, Y.; Tai, Y.; Liu, X.; Shen, C.; and Yang, J. 2018.
\newblock Fsrnet: End-to-end learning face super-resolution with facial priors.
\newblock In \emph{Proceedings of the IEEE conference on computer vision and pattern recognition}, 2492--2501.

\bibitem[{Chong et~al.(2025)Chong, Xu, Zhang, Wang, Forsyth, Krishnan, Wu, and Wang}]{chong2025copy}
Chong, M.~J.; Xu, D.; Zhang, Y.; Wang, Z.; Forsyth, D.; Krishnan, G.; Wu, Y.; and Wang, J. 2025.
\newblock Copy or Not? Reference-Based Face Image Restoration with Fine Details.
\newblock In \emph{2025 IEEE/CVF Winter Conference on Applications of Computer Vision (WACV)}, 9660--9669. IEEE.

\bibitem[{Dai et~al.(2019)Dai, Cai, Zhang, Xia, and Zhang}]{dai2019san}
Dai, T.; Cai, J.; Zhang, Y.; Xia, S.-T.; and Zhang, L. 2019.
\newblock Second-order attention network for single image super-resolution.
\newblock In \emph{Proceedings of the IEEE/CVF conference on computer vision and pattern recognition}, 11065--11074.

\bibitem[{Dai et~al.(2024)Dai, Wang, Guo, Li, Wang, and Zhu}]{dai2024freqformer}
Dai, T.; Wang, J.; Guo, H.; Li, J.; Wang, J.; and Zhu, Z. 2024.
\newblock FreqFormer: Frequency-aware transformer for lightweight image super-resolution.
\newblock In \emph{Proceedings of the International Joint Conference on Artificial Intelligence}, 731--739.

\bibitem[{Deng et~al.(2020)Deng, Guo, Ververas, Kotsia, and Zafeiriou}]{deng2020retinaface}
Deng, J.; Guo, J.; Ververas, E.; Kotsia, I.; and Zafeiriou, S. 2020.
\newblock Retinaface: Single-shot multi-level face localisation in the wild.
\newblock In \emph{Proceedings of the IEEE/CVF conference on computer vision and pattern recognition}, 5203--5212.

\bibitem[{Deng et~al.(2019)Deng, Guo, Xue, and Zafeiriou}]{deng2019arcface}
Deng, J.; Guo, J.; Xue, N.; and Zafeiriou, S. 2019.
\newblock Arcface: Additive angular margin loss for deep face recognition.
\newblock In \emph{Proceedings of the IEEE/CVF conference on computer vision and pattern recognition}, 4690--4699.

\bibitem[{Dhariwal and Nichol(2021)}]{dhariwal2021diffusion}
Dhariwal, P.; and Nichol, A. 2021.
\newblock Diffusion models beat gans on image synthesis.
\newblock \emph{Advances in neural information processing systems}, 34: 8780--8794.

\bibitem[{Dogan, Gu, and Timofte(2019)}]{dogan2019exemplar}
Dogan, B.; Gu, S.; and Timofte, R. 2019.
\newblock Exemplar guided face image super-resolution without facial landmarks.
\newblock In \emph{Proceedings of the IEEE/CVF conference on computer vision and pattern recognition workshops}, 0--0.

\bibitem[{Guo et~al.(2025)Guo, Guo, Zha, Zhang, Li, Dai, Xia, and Li}]{guo2025mambairv2}
Guo, H.; Guo, Y.; Zha, Y.; Zhang, Y.; Li, W.; Dai, T.; Xia, S.-T.; and Li, Y. 2025.
\newblock Mambairv2: Attentive state space restoration.
\newblock In \emph{Proceedings of the Computer Vision and Pattern Recognition Conference}, 28124--28133.

\bibitem[{Guo et~al.(2024)Guo, Li, Dai, Ouyang, Ren, and Xia}]{guo2024mambair}
Guo, H.; Li, J.; Dai, T.; Ouyang, Z.; Ren, X.; and Xia, S.-T. 2024.
\newblock Mambair: A simple baseline for image restoration with state-space model.
\newblock In \emph{European conference on computer vision}, 222--241. Springer.

\bibitem[{He et~al.(2022)He, Shi, Chen, Fu, and Dong}]{he2022gcfsr}
He, J.; Shi, W.; Chen, K.; Fu, L.; and Dong, C. 2022.
\newblock Gcfsr: a generative and controllable face super resolution method without facial and gan priors.
\newblock In \emph{Proceedings of the IEEE/CVF conference on computer vision and pattern recognition}, 1889--1898.

\bibitem[{Heusel et~al.(2017)Heusel, Ramsauer, Unterthiner, Nessler, and Hochreiter}]{heusel2017gans}
Heusel, M.; Ramsauer, H.; Unterthiner, T.; Nessler, B.; and Hochreiter, S. 2017.
\newblock Gans trained by a two time-scale update rule converge to a local nash equilibrium.
\newblock \emph{Advances in neural information processing systems}, 30.

\bibitem[{Ho, Jain, and Abbeel(2020)}]{ddpm}
Ho, J.; Jain, A.; and Abbeel, P. 2020.
\newblock Denoising diffusion probabilistic models.
\newblock \emph{Advances in neural information processing systems}, 33: 6840--6851.

\bibitem[{Huang et~al.(2017)Huang, He, Sun, and Tan}]{huang2017wavelet}
Huang, H.; He, R.; Sun, Z.; and Tan, T. 2017.
\newblock Wavelet-srnet: A wavelet-based cnn for multi-scale face super resolution.
\newblock In \emph{Proceedings of the IEEE international conference on computer vision}, 1689--1697.

\bibitem[{Jiang et~al.(2021)Jiang, Wang, Liu, and Ma}]{jiang2021deep}
Jiang, J.; Wang, C.; Liu, X.; and Ma, J. 2021.
\newblock Deep learning-based face super-resolution: A survey.
\newblock \emph{ACM Computing Surveys (CSUR)}, 55(1): 1--36.

\bibitem[{Karras, Laine, and Aila(2019)}]{ffhq}
Karras, T.; Laine, S.; and Aila, T. 2019.
\newblock A style-based generator architecture for generative adversarial networks.
\newblock In \emph{Proceedings of the IEEE/CVF conference on computer vision and pattern recognition}, 4401--4410.

\bibitem[{Ke et~al.(2021)Ke, Wang, Wang, Milanfar, and Yang}]{ke2021musiq}
Ke, J.; Wang, Q.; Wang, Y.; Milanfar, P.; and Yang, F. 2021.
\newblock Musiq: Multi-scale image quality transformer.
\newblock In \emph{Proceedings of the IEEE/CVF international conference on computer vision}, 5148--5157.

\bibitem[{Li et~al.(2022)Li, Yang, Chang, Chen, Feng, Xu, Li, and Chen}]{li2022srdiff}
Li, H.; Yang, Y.; Chang, M.; Chen, S.; Feng, H.; Xu, Z.; Li, Q.; and Chen, Y. 2022.
\newblock Srdiff: Single image super-resolution with diffusion probabilistic models.
\newblock \emph{Neurocomputing}, 479: 47--59.

\bibitem[{Li et~al.(2020)Li, Li, Ren, Zhang, Wang, and Zuo}]{ASFFNet}
Li, X.; Li, W.; Ren, D.; Zhang, H.; Wang, M.; and Zuo, W. 2020.
\newblock Enhanced blind face restoration with multi-exemplar images and adaptive spatial feature fusion.
\newblock In \emph{Proceedings of the IEEE/CVF Conference on Computer Vision and Pattern Recognition}, 2706--2715.

\bibitem[{Li et~al.(2018)Li, Liu, Ye, Zuo, Lin, and Yang}]{li2018learning}
Li, X.; Liu, M.; Ye, Y.; Zuo, W.; Lin, L.; and Yang, R. 2018.
\newblock Learning warped guidance for blind face restoration.
\newblock In \emph{Proceedings of the European conference on computer vision (ECCV)}, 272--289.

\bibitem[{Nitzan et~al.(2022)Nitzan, Aberman, He, Liba, Yarom, Gandelsman, Mosseri, Pritch, and Cohen-Or}]{MyStyle}
Nitzan, Y.; Aberman, K.; He, Q.; Liba, O.; Yarom, M.; Gandelsman, Y.; Mosseri, I.; Pritch, Y.; and Cohen-Or, D. 2022.
\newblock Mystyle: A personalized generative prior.
\newblock \emph{ACM Transactions on Graphics (TOG)}, 41(6): 1--10.

\bibitem[{Preechakul et~al.(2022)Preechakul, Chatthee, Wizadwongsa, and Suwajanakorn}]{preechakul2022diffusion}
Preechakul, K.; Chatthee, N.; Wizadwongsa, S.; and Suwajanakorn, S. 2022.
\newblock Diffusion autoencoders: Toward a meaningful and decodable representation.
\newblock In \emph{Proceedings of the IEEE/CVF conference on computer vision and pattern recognition}, 10619--10629.

\bibitem[{Rombach et~al.(2022)Rombach, Blattmann, Lorenz, Esser, and Ommer}]{rombach2022high}
Rombach, R.; Blattmann, A.; Lorenz, D.; Esser, P.; and Ommer, B. 2022.
\newblock High-resolution image synthesis with latent diffusion models.
\newblock In \emph{Proceedings of the IEEE/CVF conference on computer vision and pattern recognition}, 10684--10695.

\bibitem[{Saharia et~al.(2022)Saharia, Ho, Chan, Salimans, Fleet, and Norouzi}]{saharia2022image}
Saharia, C.; Ho, J.; Chan, W.; Salimans, T.; Fleet, D.~J.; and Norouzi, M. 2022.
\newblock Image super-resolution via iterative refinement.
\newblock \emph{IEEE transactions on pattern analysis and machine intelligence}, 45(4): 4713--4726.

\bibitem[{Serengil and Ozpinar(2020)}]{serengil2020lightface}
Serengil, S.~I.; and Ozpinar, A. 2020.
\newblock Lightface: A hybrid deep face recognition framework.
\newblock In \emph{2020 innovations in intelligent systems and applications conference (ASYU)}, 1--5. IEEE.

\bibitem[{Tao et~al.(2024)Tao, Gu, Zhang, Wang, and Cheng}]{tao2024overcoming}
Tao, K.; Gu, J.; Zhang, Y.; Wang, X.; and Cheng, N. 2024.
\newblock Overcoming false illusions in real-world face restoration with multi-modal guided diffusion model.
\newblock \emph{arXiv preprint arXiv:2410.04161}.

\bibitem[{Tomar et~al.(2023)Tomar, Arya, Rajput, and Rodriguez}]{tomar2023comprehensive}
Tomar, A.~S.; Arya, K.; Rajput, S.~S.; and Rodriguez, C.~R. 2023.
\newblock Comprehensive survey of face super-resolution techniques.
\newblock \emph{Digital Image Enhancement and Reconstruction}, 213--233.

\bibitem[{Wang, Chan, and Loy(2023)}]{wang2023exploring}
Wang, J.; Chan, K.~C.; and Loy, C.~C. 2023.
\newblock Exploring clip for assessing the look and feel of images.
\newblock In \emph{Proceedings of the AAAI conference on artificial intelligence}, volume~37, 2555--2563.

\bibitem[{Wang et~al.(2023)Wang, Zhang, Zhang, Zheng, Zhou, Zhang, and Wang}]{dr2}
Wang, Z.; Zhang, Z.; Zhang, X.; Zheng, H.; Zhou, M.; Zhang, Y.; and Wang, Y. 2023.
\newblock Dr2: Diffusion-based robust degradation remover for blind face restoration.
\newblock In \emph{Proceedings of the IEEE/CVF Conference on Computer Vision and Pattern Recognition}, 1704--1713.

\bibitem[{Wu and He(2018)}]{wu2018group}
Wu, Y.; and He, K. 2018.
\newblock Group normalization.
\newblock In \emph{Proceedings of the European conference on computer vision (ECCV)}, 3--19.

\bibitem[{Yang et~al.(2025)Yang, Dai, Zhu, Li, Li, and Xia}]{yang2025diffusion}
Yang, J.; Dai, T.; Zhu, Y.; Li, N.; Li, J.; and Xia, S.-T. 2025.
\newblock Diffusion Prior Interpolation for Flexibility Real-World Face Super-Resolution.
\newblock In \emph{Proceedings of the AAAI Conference on Artificial Intelligence}, volume~39, 9211--9219.

\bibitem[{Yang et~al.(2023)Yang, Zhou, Tao, and Loy}]{pgdiff}
Yang, P.; Zhou, S.; Tao, Q.; and Loy, C.~C. 2023.
\newblock PGDiff: Guiding diffusion models for versatile face restoration via partial guidance.
\newblock \emph{Advances in Neural Information Processing Systems}, 36: 32194--32214.

\bibitem[{Yue and Loy(2024)}]{difface}
Yue, Z.; and Loy, C.~C. 2024.
\newblock Difface: Blind face restoration with diffused error contraction.
\newblock \emph{IEEE Transactions on Pattern Analysis and Machine Intelligence}.

\bibitem[{Zhang et~al.(2018{\natexlab{a}})Zhang, Isola, Efros, Shechtman, and Wang}]{zhang2018unreasonable}
Zhang, R.; Isola, P.; Efros, A.~A.; Shechtman, E.; and Wang, O. 2018{\natexlab{a}}.
\newblock The unreasonable effectiveness of deep features as a perceptual metric.
\newblock In \emph{Proceedings of the IEEE conference on computer vision and pattern recognition}, 586--595.

\bibitem[{Zhang et~al.(2018{\natexlab{b}})Zhang, Li, Li, Wang, Zhong, and Fu}]{zhang2018rcan}
Zhang, Y.; Li, K.; Li, K.; Wang, L.; Zhong, B.; and Fu, Y. 2018{\natexlab{b}}.
\newblock Image super-resolution using very deep residual channel attention networks.
\newblock In \emph{Proceedings of the European conference on computer vision (ECCV)}, 286--301.

\bibitem[{Zhang, Wu, and Chen(2020)}]{zhang2020msfsr}
Zhang, Y.; Wu, Y.; and Chen, L. 2020.
\newblock MSFSR: A multi-stage face super-resolution with accurate facial representation via enhanced facial boundaries.
\newblock In \emph{Proceedings of the IEEE/CVF conference on computer vision and pattern recognition workshops}, 504--505.

\bibitem[{Zhao et~al.(2023)Zhao, Zhang, Zhong, and Shen}]{DMDNet}
Zhao, T.; Zhang, G.; Zhong, P.; and Shen, Z. 2023.
\newblock DMDnet: A decoupled multi-scale discriminant model for cross-domain fish detection.
\newblock \emph{Biosystems Engineering}, 234: 32--45.

\bibitem[{Zhou et~al.(2022)Zhou, Chan, Li, and Loy}]{codeformer}
Zhou, S.; Chan, K.; Li, C.; and Loy, C.~C. 2022.
\newblock Towards robust blind face restoration with codebook lookup transformer.
\newblock \emph{Advances in Neural Information Processing Systems}, 35: 30599--30611.

\end{thebibliography}

\end{document}